\documentclass[letterpaper, 10 pt, conference]{ieeeconf}  % Comment this line out if you need a4paper

\IEEEoverridecommandlockouts                              % This command is only needed if 
\usepackage{amsmath} % assumes amsmath package installed
\usepackage{amssymb}  % assumes amsmath package installed
\usepackage{subfig}	% for subfigures
\usepackage{booktabs}
\usepackage{url}
\usepackage{multirow}
\usepackage{siunitx}
\usepackage{todonotes}
\definecolor{est}{rgb}{0.25,0.4,0.9}
\definecolor{conf}{rgb}{0.6,0.4,0}

\DeclareMathOperator*{\argmin}{\arg\min} 

\title{\LARGE \bf A Modular Approach to the Embodiment of Hand Motions from Human Demonstrations}

\author{Alexander Fabisch$^{1}$, Manuela Uliano$^2$, Dennis Marschner$^{3}$, Melvin Laux$^{3}$, Johannes Brust$^4$, Marco Controzzi$^2$% <-this % stops a space
\thanks{This work was supported by the European Commission under the Horizon 2020 framework program for Research and Innovation (project acronym: APRIL, project number: 870142).}% <-this % stops a space
\thanks{$^{1}$
        Robotics Innovation Center, DFKI GmbH,
        Robert-Hooke-Stra{\ss}e 1, D-28359 Bremen, Germany
        ({\tt\small alexander.fabisch@dfki.de})}%
\thanks{$^{2}$ The BioRobotics Institute, Scuola Superiore Sant'Anna, Pisa, Italy, and with the Department of Excellence in Robotics \& AI, Pisa, Scuola Superiore Sant'Anna, Italy. }%
\thanks{$^{3}$
        Robotics Research Group, University of Bremen %,
        }%
\thanks{$^{4}$
        Plan-Based Robot Control, DFKI GmbH %,
        }
}

\begin{document}

\maketitle
\thispagestyle{empty}
\pagestyle{empty}

%%%%%%%%%%%%%%%%%%%%%%%%%%%%%%%%%%%%%%%%%%%%%%%%%%%%%%%%%%%%%%%%%%%%%%%%%%%%%%%%
\begin{abstract}
Manipulating objects with robotic hands is a complicated task.
Not only the fingers of the hand, but also the pose of the robot's end effector
need to be coordinated.
Using human demonstrations of movements is an intuitive and data-efficient
way of guiding the robot's behavior.
We propose a modular framework with an automatic embodiment mapping to
transfer recorded human hand motions to robotic systems.
In this work, we use motion capture to record human motion.
We evaluate our approach on eight challenging tasks, in which a robotic
hand needs to grasp and manipulate either deformable or small and fragile
objects. We test a subset of trajectories in simulation and on a real
robot and the overall success rates are aligned.

\end{abstract}

%%%%%%%%%%%%%%%%%%%%%%%%%%%%%%%%%%%%%%%%%%%%%%%%%%%%%%%%%%%%%%%%%%%%%%%%%%%%%%%%
\section{INTRODUCTION}
Although manipulation of known objects is a well-studied field, handling deformable or small, fragile objects with human-level skill is a challenge. Behaviors for robotic hands can be generated through various approaches, e.g., planning, reinforcement learning, or imitation learning. %\cite{Billard2008,Argall2009,Osa2018}.
We are interested in leveraging intuitive human knowledge to generate data for imitation learning with a complex hand. Dataset generation is difficult in this case. Kinesthetic teaching becomes tricky when a 5-finger hand and the end effector's pose need to be controlled. Teleoperation might not exploit the full potential of the human demonstration due to restricted movement or control difficulties.
We propose to use external sensors (motion capture) to track human hands and transfer their states to robotic hands. To do this, we infer the human hand's state with a record mapping \cite{Argall2009}.  Next, we solve the correspondence problem \cite{Nehaniv2002}, induced by kinematic differences between human and robotic hands, with an embodiment mapping \cite{Argall2009}.
%Methods like reinforcement learning can be used to refine transferred motions.

\begin{figure}[tpb]
\centering
\includegraphics[width=0.95\columnwidth]{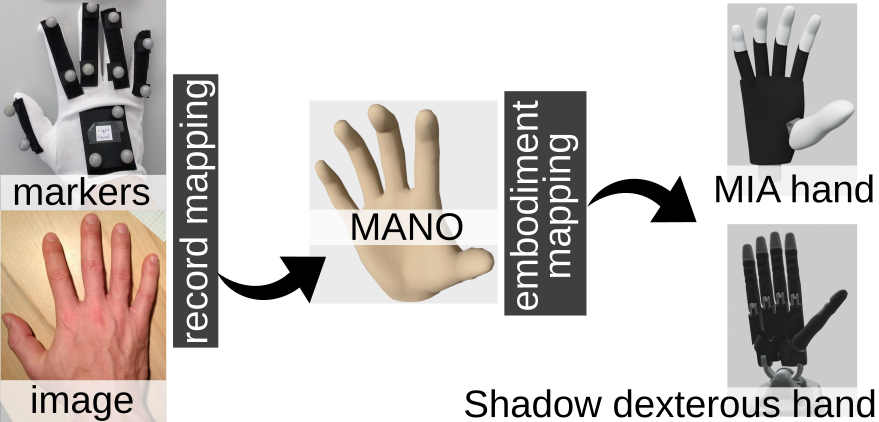}
\caption{Proposed approach to embodiment of hand motions.}
\label{fig:overview}
\vspace{-1.5em}
\end{figure}

Our goal is to develop a modular framework that allows us to easily
replace the sensor as well as the target system (see
Figure~\ref{fig:overview}). For example, switching between different optical
methods, e.g., motion capture and camera-based hand tracking, should be easy.
For this reason, we use the MANO
hand model \cite{Romero2017}, which has previously been used in
camera-based hand tracking \cite{Hasson2019}, as an intermediate
representation of the hand's state. The embodiment mapping should also be
configurable to handle multiple robotic hands.
The implementation is available at
\url{https://github.com/dfki-ric/hand_embodiment}.

\section{BACKGROUND AND RELATED WORK}

\subsection{Motion Capture of Human Hands}
Capturing human hand motions as fully articulated 3D hand poses is demanding
due to the dexterity of hands and high angular velocities.
Nevertheless, the task is well studied and there are numerous solutions,
including optical, non-optical, and hybrid methods.

\subsubsection{Non-optical Methods}
Methods based on electromagnetic transmitters \cite{Shen2021,Ma2011,Chen2016},
bending \cite{Gentner2009,Shen2016,Ciotti2016ASO} or stretch-sensors
\cite{Chossat2015,Atalay2017,Glauser2019}, inertial measurement units
\cite{Mankowski2017,Hsiao2015,Connolly2015}, or even exoskeletons
\cite{Pereira2019} are often integrated as gloves. Caeiro-Rodríguez~et~al.
provide a review of commercial active smart gloves \cite{CaeiroRodriguez2021}.
These methods are suitable for real-time applications, but several problems,
such as complex calibration and noisy data with drift over time, remain.
Considering different hand shapes, it is not trivial to place multiple sensors
perfectly on the glove without loss of accuracy or hand shape-dependent
calibration methods.

\subsubsection{Optical Methods}
The continuum of optical methods ranges from estimators based on markerless,
monocular color images to marker-based motion capture systems using multiple
cameras.
Markerless methods, mostly based on deep learning, led to groundbreaking
progress in computer vision.
However, various factors such as lighting conditions, image resolution,
background and skin color can influence their performance.
Optical markerless approaches can be divided into generative
\cite{Spurr2018,Gu2020} and discriminative methods
\cite{Zimmermann2017,Mueller2018,Panteleris2017}.
%Generative methods attempt to learn a manifold of hand poses in a latent space via a combination of variational autoencoders (VAEs) and generative adversarial networks (GANs).
%In the discriminative 2D hand and finger keypoints are first obtained in form of heat maps or probability maps via convolutional neural networks (CNNs). Subsequently, 3D pose joints are estimated from these labeled keypoints via a regression network or by fitting a 3D hand model.
Depth information can improve the accuracy of markerless optical methods.
However, most methods reach their limits in everyday applications because
they generalize insufficiently. Occlusion and self-occlusion when interacting
with objects are problems that can be addressed with multi-view approaches
\cite{Sridhar2013,Tzionas2015,Wang2011,Simon2017}.
%Shanxin Yuan et al. analyzed the top ten among seventeen participating methods from the HIM2017 challenge \cite{Yuan2017HMC} to summarize the current state-of-the-art of 3D hand pose estimation from depth images \cite{Yuan2017HPE}. Their findings include, that current methods perform well on single hand pose estimation when trained on a million-scale dataset, but have difficulties in generalizing to hand-object interaction.
%Simon et al. proposed an approach called "multiview bootstrapping" that uses a multi-camera system to improve 2D hand keypoint detectors, which are weak in the case of occlusion \cite{Simon2017}. By triangulating over multiple views, the system can iteratively correct misjudgments. As a result, the improved single view detector estimates better 3D hand poses even in the presence of complex object interactions.

%The same multi-view procedure is commonly known from professional marker-based motion capture (MOCAP) systems.
Optical marker-based methods (motion capture, MOCAP) are
widely used in both the film industry to create realistic animations
\cite{Zhang2013} and for motion analysis in sports biomechanics and
rehabilitation \cite{Colyer2018}. With proper hardware and environment,
professional MOCAP systems estimate hand poses more accurately than markerless
methods. With a growing number of perspectives and higher camera
resolutions, marker detection accuracy, as well as robustness against
occlusions increase. However, such systems are expensive and of little
use outside of laboratories.

\subsection{Human Hand Pose Models}
Cobos et al. show that 24 degrees of freedom (DOF) are suitable for modeling
the high kinematic
complexity of human hands during manipulation \cite{Cobos2008}. Yet, no
universal kinematic hand model is equally suitable for all capturing methods,
and the number of measurement points varies between different hardware
setups. In non-optical methods, labeled 3D joints of hand and finger key
points, as well as joint angles, are commonly measured. Optical approaches
usually estimate labeled 2.5D or 3D joints, but not joint angles. Mostly,
labeled 3D points are assigned to a hand skeleton, which is helpful for
advanced applications. However, human hand poses can also be represented as
differentiable 3D hand models such as MANO \cite{Romero2017}, whose surface
mesh can be fully deformed and posed. Compared to only regressing a 3D hand
skeleton, this 3D hand mesh makes the method usable for computer
vision and embodiment mapping.
%Furthermore mesh generative human hand models can be helpful for adapting color image-based 3D hand pose estimation to non-laboratory domains and additionally improving results on common benchmarks. Kulon et. Al. integrated MANO in a promising approach for end-to-end neural network training with mesh supervision that is obtained through an automated data collection method. \cite{Kulon2020}

\subsection{Embodiment Mapping}
Embodiment mappings solve the problem of fitting movements demonstrated by a
human to a robotic target system. 
The main challenge of this task is to deal with the differences between
kinematic structures and dynamics of humans and robots. 
Previous works define complex objective functions that have to be solved for a
complete trajectory, and focus on robotic arms
\cite{Maeda2016,Gutzeit2018,Gutzeit2019} that have less variety in kinematic
design than robotic hands.

We aim to design an embodiment mapping for different robotic hands and input
modalities. This is achieved by using MANO \cite{Romero2017} as an
intermediate hand state representation, i.e., as an adapter between
input modalities and target systems, while previous approaches only support
one input modality and only work with robotic arms.

% We aim to design an embodiment mapping for robotic hands with different
% properties. We consider individual steps instead of complete motions, which
% enables us to map each state of the human hand online if done fast enough.
% Hence, the objective can be decoupled into multiple independent problems:
% one for each finger per step of the motion.

\subsection{Robotic Hands}
We consider two robotic hands as target platforms: Prensilia's Mia Hand, as
an example of a simple, robust robotic hand, and the Shadow Dexterous Hand of
Shadow Robot Company as an example of a complex, fragile hand.\footnote{%
Although we mostly evaluate these two hands, the BarrettHand and the Robotiq
2F-140 gripper are already integrated in the open source release.}

\subsubsection{Mia Hand}
The Mia Hand is a simple, but robust robotic hand with 4 DOF that
can be controlled at \SI{20}{\hertz}. The controllable joints are:
thumb adduction/abduction (binary) and flexion, index finger flexion as well as
coupled flexion of middle, ring, and little finger, which are controlled by the
same motor. As it is not possible to quickly switch between adduction and
abduction of the thumb, we consider this joint to be fixed.
% It is equipped with sensors that measure normal and tangential forces at the thumb, index finger, and middle finger.

\subsubsection{Shadow Dexterous Hand}
The Shadow Dexterous Hand is complex as it has 24 DOF, of which 20 are
controlled actively at \SI{500}{\hertz}.
The last two joints of each finger (except the thumb)
are coupled, such that the last joint moves when the previous one reaches the
joint limit and vice versa. 
% It is equipped with tactile sensors at each finger tip.

\section{MODULAR RECORD AND EMBODIMENT MAPPING FOR ROBOTIC HANDS}

We propose a modular framework to transfer human hand motions to robotic
hands. Modularity allows us to easily adapt to new input modalities
and target systems.

\subsection{Desiderata}
To transfer hand motions demonstrated by humans to a robot, we define a
record mapping and an embodiment mapping.
Our approach is designed to fulfill the following criteria:
\begin{itemize}
\item The approach should be adaptable to different input modalities
by replacing the record mapping.
\item Hence, the result of the record mapping should be a common
representation of human hand states.
\item The embodiment mapping should be fast enough to enable immediate transfer
in teleoperation scenarios.
\item The embodiment mapping should be able to adapt to the target system
through configuration.
\end{itemize}

\subsection{Record Mapping for Motion Capture System}
\begin{figure}[tb]
\centering
\vskip 1em  % fix margin impositions
\includegraphics[width=0.7\columnwidth]{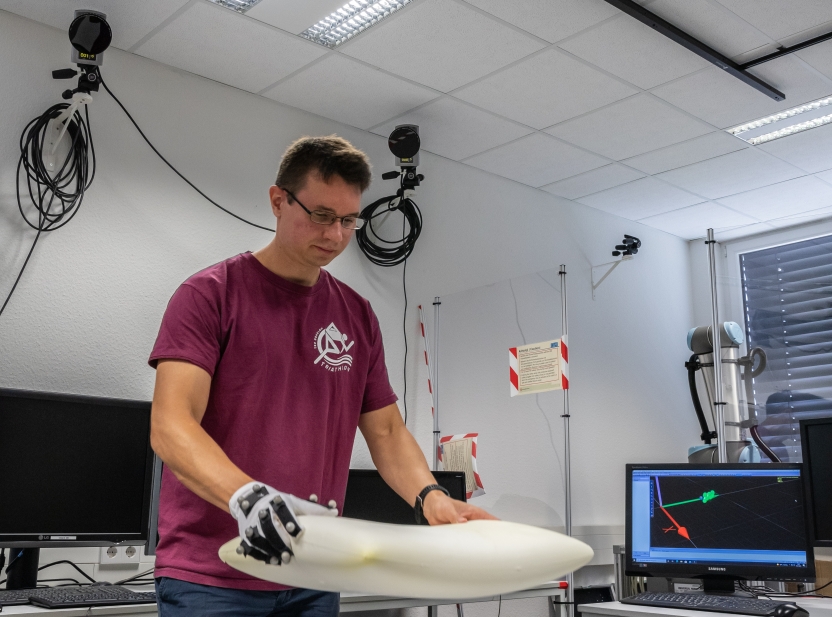}
\caption{Motion capture experiment.}
\label{fig:mocap}
\vskip -1em
\end{figure}

\begin{figure}[thpb]
\centering
\hfill
\subfloat[Motion capture glove with frame defined by markers at the back of the hand.]{%
\includegraphics[width=1.4in]{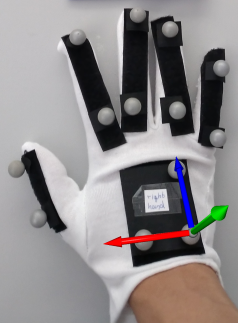}\label{fig:mocap_markers}}%
\hfill
\subfloat[MANO model with expected marker positions indicated by green spheres.]{%
\includegraphics[width=1.4in]{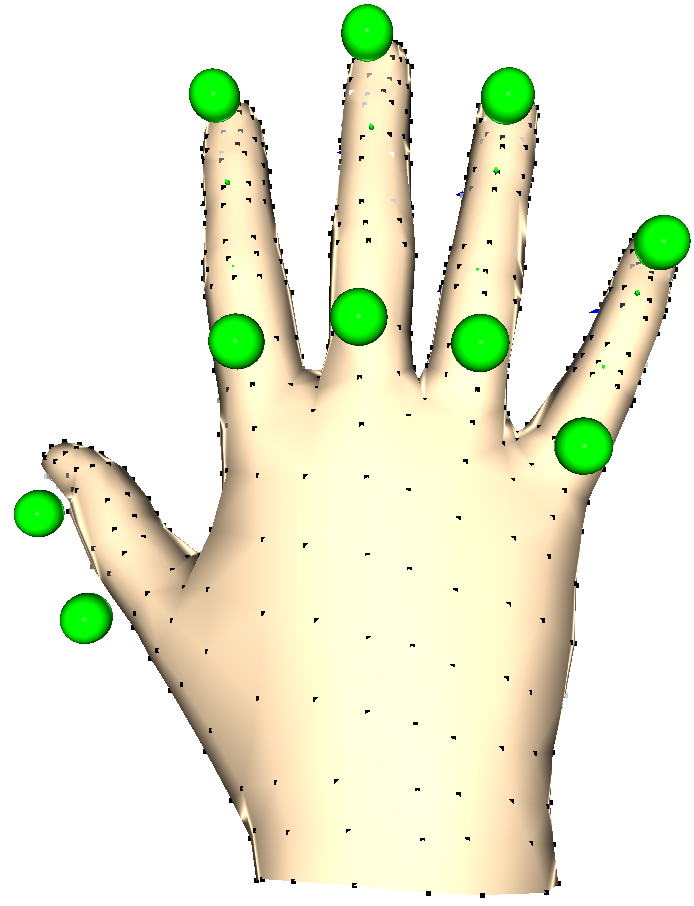}\label{fig:mano_markers}}
\hfill
\caption{Mapping from motion capture markers to MANO.}
\label{fig:markers}
\end{figure}

% https://git.hb.dfki.de/dfki-interaction/experimental/hand_embodiment/-/blob/master/hand_embodiment/record_markers.py
The objective of the record mapping is to estimate the state of the MANO
model from motion capture markers. We use a Qualisys MOCAP system and a glove
with 13 passive markers (see Figure \ref{fig:mocap})
for the right hand.
Three markers on the back of the hand and two markers per finger are used
to reconstruct the pose of the hand and the configuration of each finger
(see Figure \ref{fig:mocap_markers}).
%Three markers on the back of the hand can be translated to the hand's pose and
%two markers are used per finger (see Figure \ref{fig:mocap_markers}).

We use colors to distinguish between \textcolor{est}{estimated or measured
quantities} and \textcolor{conf}{configuration parameters} in formulas.
To estimate the pose $T_{\texttt{world},\texttt{MANO}} \in SE(3)$ (read:
active transformation from MANO frame to world frame) of the MANO model,
we first derive the pose $\textcolor{est}{T_{\texttt{world},\texttt{hand}}} \in SE(3)$
of the hand based on three labeled markers on the back of the hand.
%of the back of the hand based on three labeled markers.
In accordance with
the two-vector representation \cite{Corke2017}, we define the hand frame
orientation by the approach vector (direction from right to front hand marker)
and the orientation vector (normal of the plane defined by the
three markers). 
The origin of the hand frame can be any point in the plane of the three markers.
Our frame convention is shown in Figure \ref{fig:mocap_markers}.
%The hand's position can be any point in the plane of the three markers.
% We chose the middle between left and right hand markers.
When we know the fixed transformation
$\textcolor{conf}{T_{\texttt{hand},\texttt{MANO}}} \in SE(3)$,
we compute
\[T_{\texttt{world},\texttt{MANO}} = \textcolor{est}{T_{\texttt{world},\texttt{hand}}} \textcolor{conf}{T_{\texttt{hand},\texttt{MANO}}}.\]

With the known pose of the MANO model, estimating the finger states boils down
to solving five individual optimization problems. We compute each finger's
forward kinematics,
$f_{\textcolor{conf}{\boldsymbol{\beta}},i,j}(\boldsymbol{q}) = \boldsymbol{p}_{i,j},$
for the two points $\boldsymbol{p}_{i,1},\boldsymbol{p}_{i,2}$ (see Figure
\ref{fig:mano_markers}), where $\boldsymbol{q}_i \in \mathbb{R}^9$ are the
joint angles of finger $i \in \{1,...,5\}$.
The resulting optimization problems for each finger are defined as
\begin{eqnarray*}
\boldsymbol{q}_i^* = \argmin_{\boldsymbol{q}_i} &
\sum_{j=1}^2||\textcolor{est}{\hat{\boldsymbol{p}}_{i,j}} - f_{\textcolor{conf}{\boldsymbol{\beta}},i,j}(\boldsymbol{q}_i)||^2 + R(\boldsymbol{q}_i)\\
\text{subject to} & \textcolor{conf}{\boldsymbol{q}^{\min}_{i}} \leq \boldsymbol{q}_i \leq \textcolor{conf}{\boldsymbol{q}^{\max}_{i}},
\end{eqnarray*}
where we penalize each joint angle individually in positive and negative direction with
$R(\boldsymbol{q}_i) = ||\max(\textcolor{conf}{\boldsymbol{w}_{i,+}} \circ \boldsymbol{q}_i, \boldsymbol{0})||^2 + ||\min(\textcolor{conf}{\boldsymbol{w}_{i,-}} \circ \boldsymbol{q}_i, \boldsymbol{0})||^2$
with weights $\textcolor{conf}{\boldsymbol{w}_{i,+}},\textcolor{conf}{\boldsymbol{w}_{i,-}} \in \mathbb{R}^9$
($\circ$ is the Hadamard product and $\min$, $\max$ are element-wise operators).
$\textcolor{conf}{\boldsymbol{q}^{\min}_{i}},\textcolor{conf}{\boldsymbol{q}^{\max}_{i}} \in \mathbb{R}^9$ are
lower and upper bounds for joint angles, and $\textcolor{conf}{\boldsymbol{\beta}} \in \mathbb{R}^{10}$
are shape parameters of the MANO model. $\textcolor{est}{\hat{\boldsymbol{p}}_{i,j}}$
are measured positions of motion capture markers. We solve these optimization problems
with sequential least squares programming (SLSQP, \cite{Kraft1988})
and numerically estimated gradients.

%$\boldsymbol{q}_i^* \in \mathbb{R}^9$ (for $i = 1, \ldots, 5$) and $T_{\texttt{world},\texttt{MANO}} \in SE(3)$ define the full state of the MANO model, from which we can compute the corresponding marker points $\boldsymbol{p}_{i,j}^* \in \mathbb{R}^3$ (for $i = 1, \ldots, 5$ and $j = 1, 2$).
The MANO model's full state is defined by $\boldsymbol{q}_i^* \in \mathbb{R}^9$ and $T_{\texttt{world},\texttt{MANO}} \in SE(3)$, from which we can compute the corresponding marker points $\boldsymbol{p}_{i,j}^* \in \mathbb{R}^3$.

\subsection{Embodiment Mapping}

\begin{figure}[tb]
\centering
\hfill
\subfloat[Model of Mia hand with expected marker positions.]{%
\includegraphics[width=1.4in]{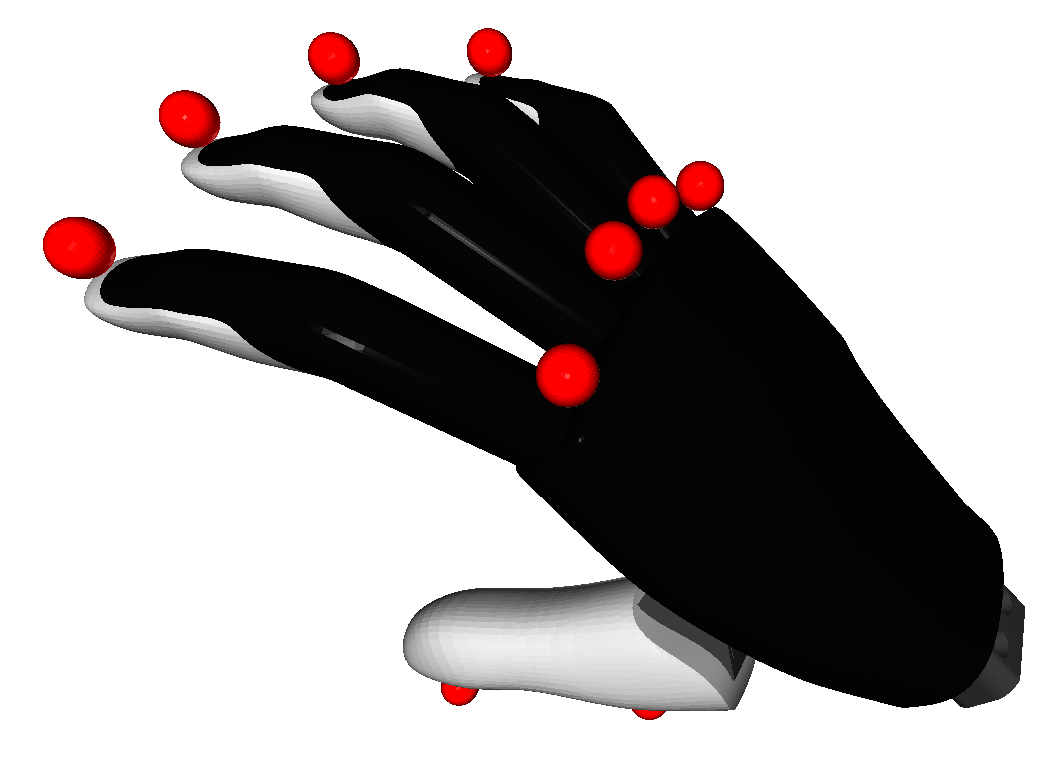}\label{fig:extended_mia}}%
\hfill
\subfloat[Model of Shadow dexterous hand with expected marker positions.]{%
\includegraphics[width=1.4in]{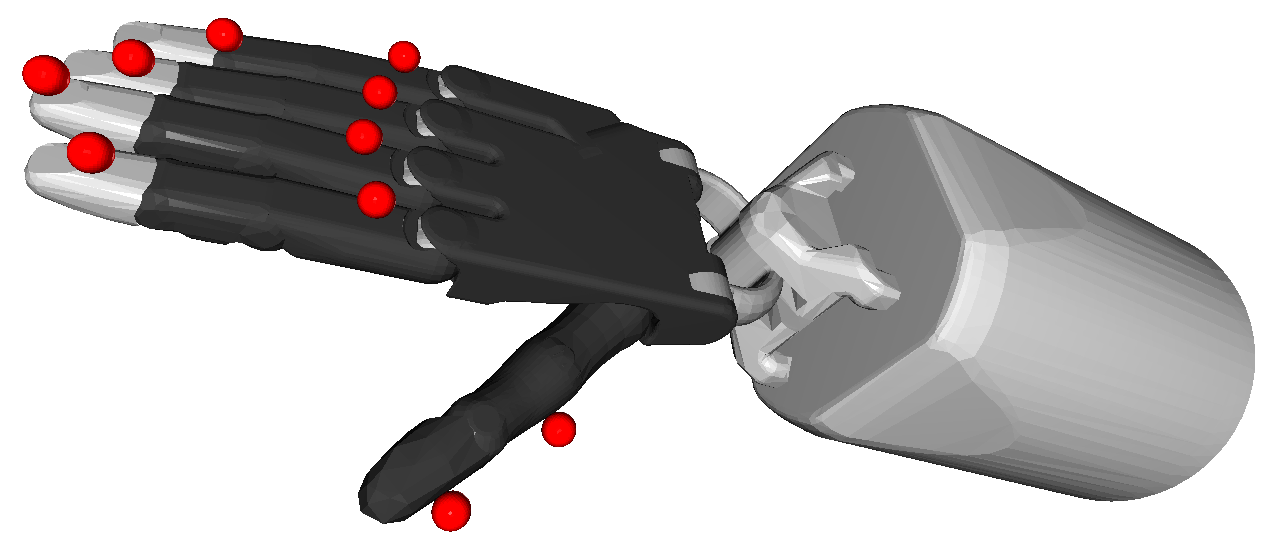}\label{fig:extended_shadow}}
\hfill
\caption{Extended kinematic hand models.}
\label{fig:kinematic}
\end{figure}

% https://git.hb.dfki.de/dfki-interaction/experimental/hand_embodiment/-/blob/master/hand_embodiment/embodiment.py
The embodiment mapping translates MANO states to states of the target system,
which is a combination of a robotic arm and hand.
Assuming that poses are reachable,
the robotic hand's pose first needs to be matched to the mesh pose, i.e.,
we must define $\textcolor{conf}{T_{\texttt{robot,MANO}}} \in SE(3)$.

Next, the individual finger configurations are optimized to be as close as
possible to the MANO mesh. 
Without real-time constraints, the ideal solution is to define an objective
function to either maximize the overlap between the volumes or to minimize
the distance between the inner surfaces of MANO's fingers and the fingers
of the robotic hand. 
With the intention to be able to transfer motions in real-time, we propose
a simplified approach. 
We define points with respect to the links of the robotic hands (see Figure
\ref{fig:kinematic}) and minimize the distance to their corresponding virtual
markers on the MANO mesh (see Figure \ref{fig:mano_markers}), for which the
positions are known from the record mapping.

Thus, the optimization of finger joints reduces to an inverse kinematics
problem, in which only the distance between two pairs of points per finger
$i$ is minimized:
\begin{equation*}
\boldsymbol{r}_i^* = \argmin_{\boldsymbol{r}_i}
\sum_{j=1}^2||f_{\textcolor{conf}{\boldsymbol{\beta}},i,j}(\boldsymbol{q}_i^*) - g_{i,j}(\boldsymbol{r}_i)||^2,
\end{equation*}
subject to $\boldsymbol{r}^{\min}_{i} \leq \boldsymbol{r}_i \leq \boldsymbol{r}^{\max}_{i}$.
$f_{\textcolor{conf}{\boldsymbol{\beta}},i,j}(\boldsymbol{q}_i^*)$
is known from record mapping and $g_{i,j}(\boldsymbol{r}_i)$ is the
corresponding forward kinematics function for the robotic hand with the joint
angles $\boldsymbol{r}_i \in \mathbb{R}^{N_i}$ and limits
$\boldsymbol{r}^{\min}_{i}, \boldsymbol{r}^{\max}_{i} \in \mathbb{R}^{N_i}$.
The number of optimized joints $N_i \in \mathbb{N}$ 
depends on the target system, as e.g., the Mia hand's index finger is
controlled by a single motor while the Shadow dexterous hand uses three
motors to control the index finger.
%(although there are four joints, two of them controlled by the same motor, which we can hide behind the abstraction of forward kinematics) 
%Once again, we use SLSQP to solve the optimization problem.
The optimization problem is solved by SLSQP.

\subsection{Configuration}
While we assume well-defined kinematics, it is necessary to configure
certain parameters of the record and embodiment mapping. For this work,
these parameters were configured manually, however, this could
be partially automated. For instance, a black-box optimizer could set
the shape parameters for the MANO model to fit motion capture markers.

For the record mapping, we need to configure:
\begin{itemize}
\item $\textcolor{conf}{T_{\texttt{hand},\texttt{MANO}}} \in SE(3)$:
transformation between MANO base and hand coordinate frame defined
by three motion capture markers at the back of the hand
\item $\textcolor{conf}{\boldsymbol{\beta}} \in \mathbb{R}^{10}$:
shape parameters of MANO
\item $\textcolor{conf}{\boldsymbol{w}_{i,+}},\textcolor{conf}{\boldsymbol{w}_{i,-}} \in \mathbb{R}^9$:
weights to penalize each joint angle individually in both directions
\item $\textcolor{conf}{\boldsymbol{q}^{\min}_{i}},\textcolor{conf}{\boldsymbol{q}^{\max}_{i}} \in \mathbb{R}^9$: joints' lower and upper bounds
\end{itemize}

For the embodiment mapping, we need to configure
\begin{itemize}
\item $\textcolor{conf}{T_{\texttt{robot,MANO}}} \in SE(3)$:
transformation between the MANO mesh's and the robotic hand's bases
\item expected marker positions (see Figure \ref{fig:kinematic})
with respect to corresponding frames in the hand's kinematic tree
\end{itemize}

\section{EVALUATION}

\subsection{Research Question}
It has been shown that the state of the MANO model can be obtained from RGB
images \cite{Hasson2019}. Our goal is to evaluate whether
\begin{enumerate}
\item It is possible to obtain the MANO representation from motion capture
data.
\item The embodiment mapping can be adapted to both robotic hands.
\item The embodiment mapping obtains plausible configurations
of the robotic hand even when the target system has less DOF than a human hand.
\item The transferred trajectories result in physically verified
useful behaviors of the target systems.
\item Both record and embodiment mapping can be executed at a frequency
suitable for teleoperation.
\end{enumerate}

\subsection{Datasets}

\begin{table}[tb]
\centering
\vskip 1em  % fix margin impositions
\begin{tabular}{rlll}
\toprule
No. & Task & Variations & Demonstrations\\
\midrule
1 & Grasp insole & from front or back & 213\\
2 & Insert insole & - & 12\\
3 & Grasp small pillow & from four sides & 224\\
4 & Grasp big pillow & from four sides & 130\\
5 & Grasp electronic component & from all sides & 55\\
6 & Assemble electronic components & from all directions & 54\\
7 & Flip pages & - & 38\\
8 & Insert passport in box & - & 37\\
\midrule
& Total & & 763\\
\bottomrule
\end{tabular}
\caption{Overview of datasets used for evaluation.}
\label{tab:datasets}
\vskip -1em
\end{table}

\begin{figure}[tb]
\centering
\includegraphics[width=0.8\columnwidth]{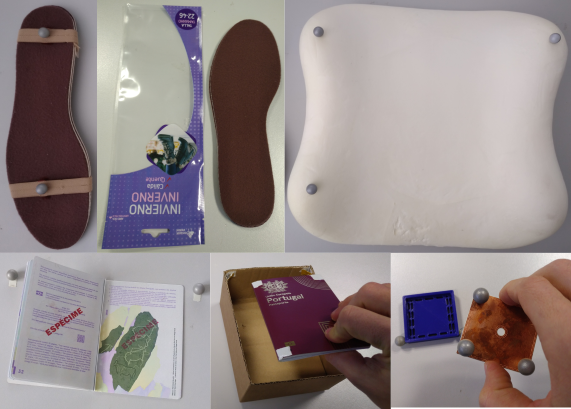}\\
\caption{Objects used to record datasets.
Left to right and top to bottom: insole with markers,
insole and bag, small pillow with markers, open passport,
passport and box, electronic components with markers.}
\label{fig:objects}
\vskip -1em
\end{figure}

\begin{figure}[tb]
\centering
\vskip 1em  % fix margin impositions
\subfloat[Dataset of 224 grasps for a small pillow.]{%
\includegraphics[width=2.3in]{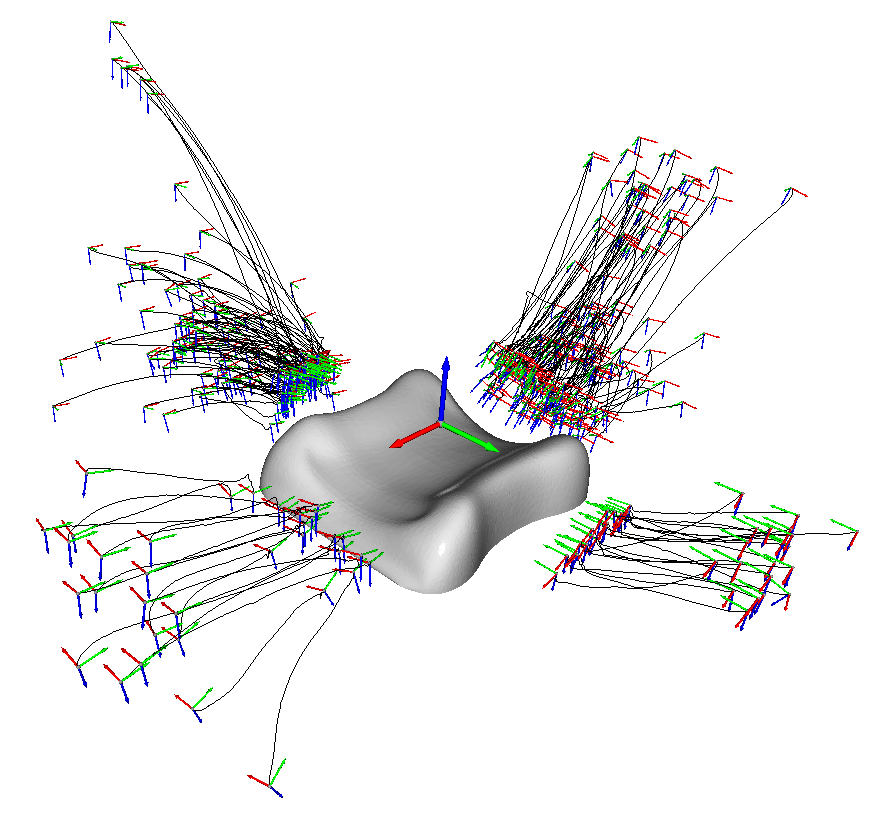}\label{fig:dataset_pillow}}\\
\subfloat[Dataset of 213 grasps for an insole.]{%
\includegraphics[width=2.8in]{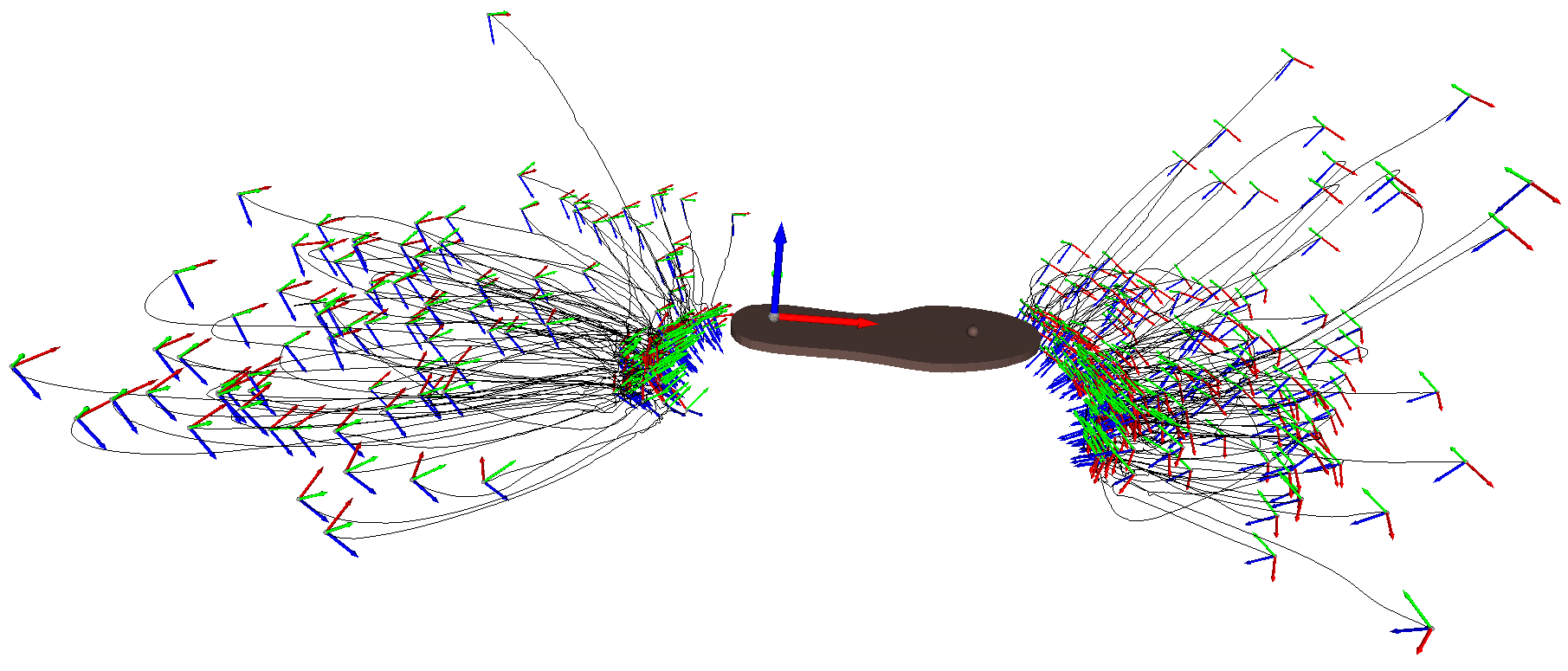}\label{fig:dataset_insole}}
\caption{Two of the datasets used for evaluation. Object-relative trajectories
of the end effector are represented by lines and coordinate frames that
indicate the orientation at the beginning and end. The large
coordinate frames in the middle define object poses.}
\label{fig:datasets}
\vskip -1.5em
\end{figure}

To evaluate the hand embodiment mapping, we recorded demonstrations of
multiple tasks and variations of these with a Qualisys MOCAP system.
Table \ref{tab:datasets} describes the tasks and reports the number of
demonstrations for each of these.\footnote{Only
one subject was recorded because of COVID-19. We argue that this is sufficient
since parameters of the record mapping are tuned manually.}
Objects that were used during these experiments
are shown in Figure \ref{fig:objects} and visualizations of the
end-effector trajectories in Figure \ref{fig:datasets}.

\subsection{Estimation of MANO State (Qualitative Evaluation)}

\begin{figure}[thpb]
\centering
\hfill
\subfloat[Grasping and assembling electronic components.]{%
\includegraphics[height=1.7in]{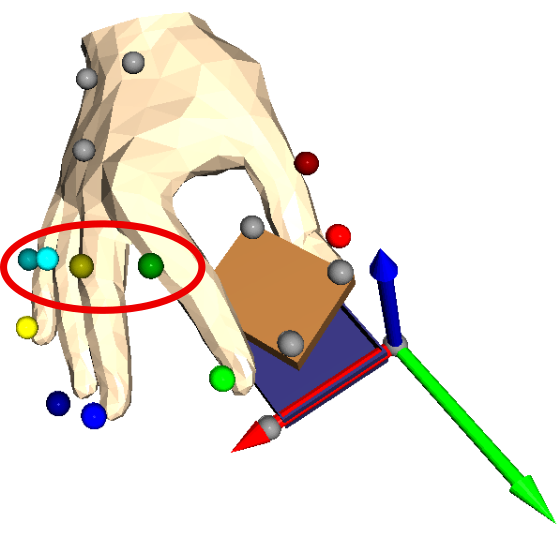}
\includegraphics[height=1.7in]{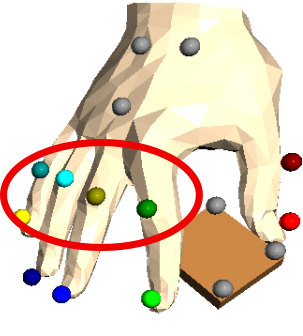}\label{fig:record_1}}%
\hfill
\subfloat[Grasping a small pillow.]{%
\includegraphics[height=1.15in]{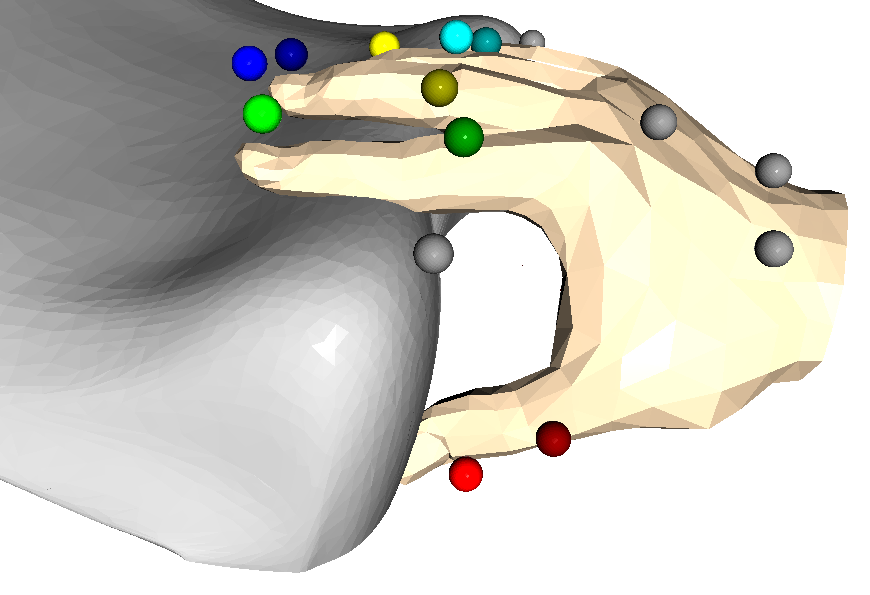}
\includegraphics[height=1.15in]{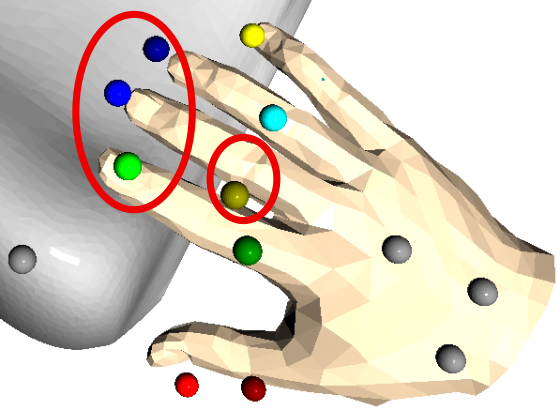}\label{fig:record_3}}%
\hfill
\subfloat[Grasping an insole.]{%
\includegraphics[height=1.0in]{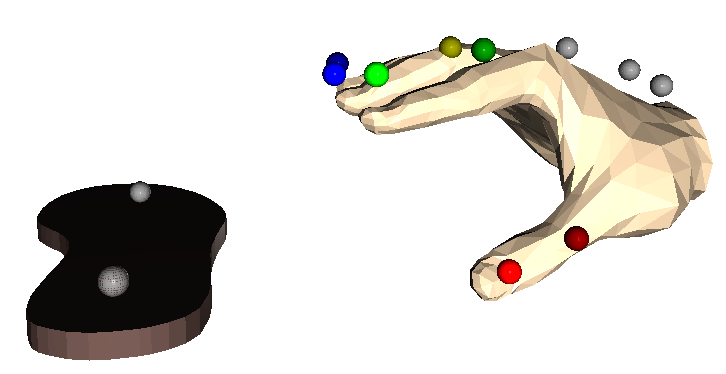}
\includegraphics[height=1.0in]{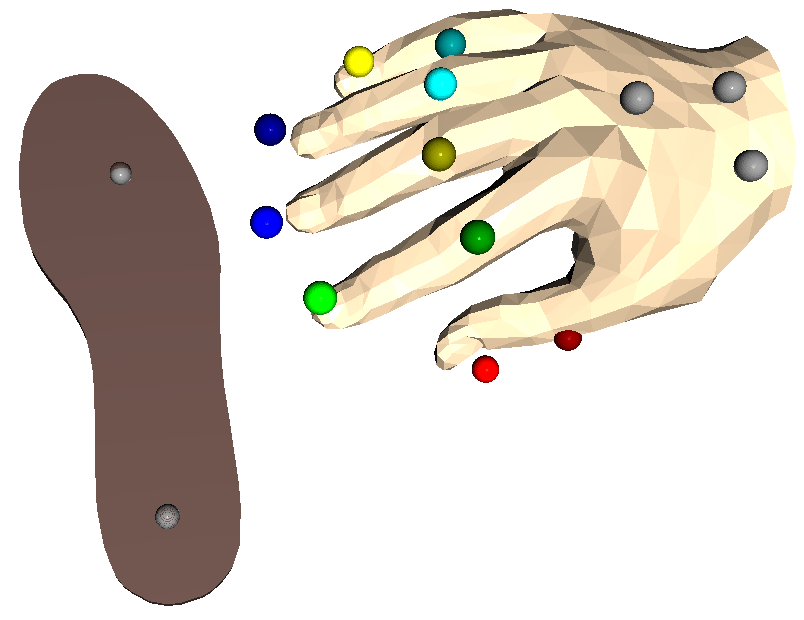}\label{fig:record_5}}%
\hfill
\subfloat[Flipping the page of a passport.]{%
\includegraphics[height=1.5in]{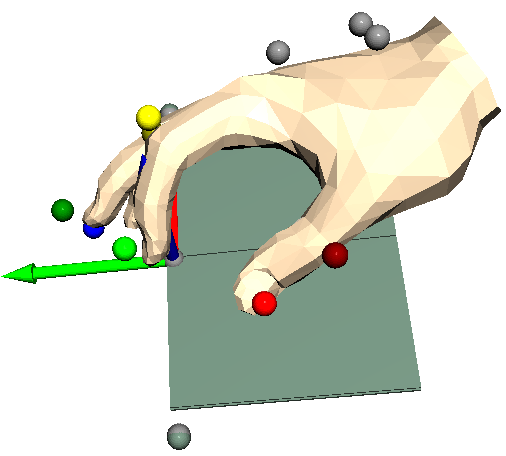}
\includegraphics[height=1.5in]{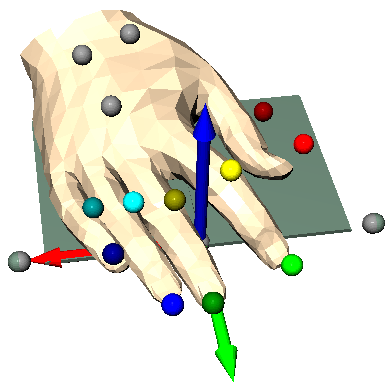}\label{fig:record_7}}%
\hfill
\subfloat[Grasping a passport and putting it in a box.]{%
\includegraphics[height=1.5in]{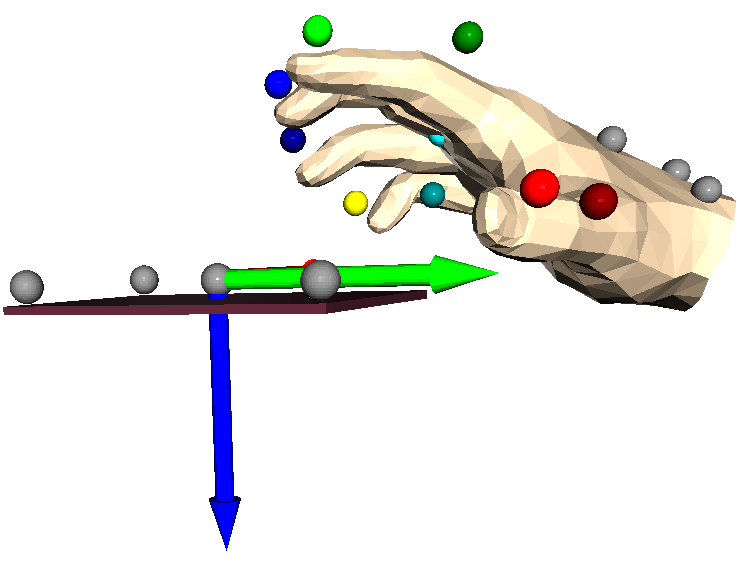}
\includegraphics[height=1.5in]{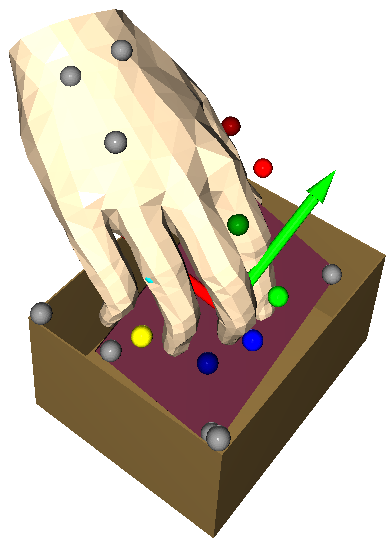}\label{fig:record_9}}%
\hfill
\caption{Exemplary configurations of MANO mesh after record mapping
and corresponding motion capture markers.
Simplified meshes that illustrate the position of the manipulated
objects are displayed with the markers that we used to track their
pose. For some objects we also see the object frame.
Illustrations were made with Open3D \cite{Zhou2018}.}
\label{fig:record_mapping}
\end{figure}

Marker positions were tracked with an error of about \SI{1}{\milli\meter}.
We can also use them to evaluate the quality of the estimated MANO states.
Figure \ref{fig:record_mapping} shows exemplary measurements
of the motion capture markers with corresponding estimations of the
MANO model by the record mapping from marker positions.
Differences between the MANO state and the actual hand mainly stem from
inaccuracies of the MANO configuration.
In particular, shape and the placement of the three markers at the back of the
MANO model are different. Marker
placement also varies between experiments and even within individual
recordings.

Obvious differences between the estimated MANO state and the actual hand
state can be seen, e.g., in Figure \ref{fig:record_1} (red ellipses): while the marker positions
are closer to the metacarpophalangeal joint of the real hand
(see Figure \ref{fig:mocap_markers}), they are closer to the proximal interphalangeal
joint of the estimated MANO state.
Furthermore, we can see in Figure \ref{fig:record_3} (red ellipses) that the lengths of the
fingers do not always match the corresponding marker positions. In the same
example, the marker close to the metacarpophalangeal joint of the middle
finger is laterally shifted, which was due to the marker not being perfectly
aligned at the center of the finger.
Nevertheless, we can see in Figure \ref{fig:record_mapping} that the
estimated MANO states are generally plausible explanations of the
measured marker positions.

\subsection{Adaptability to Robotic Hand (Qualitative Evaluation)}
\begin{figure}[hb]
\centering
\hfill
\subfloat[]{%
\includegraphics[width=0.7in]{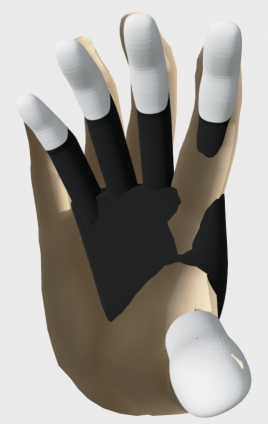}\label{fig:embodiment_interactive_mia_1}}%
\hfill
\subfloat[]{%
\includegraphics[width=0.7in]{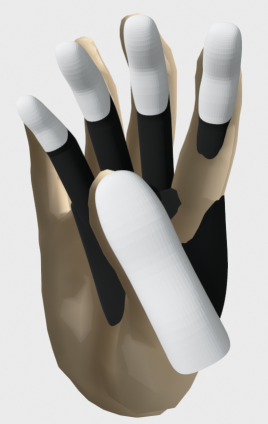}\label{fig:embodiment_interactive_mia_2}}%
\hfill
\subfloat[]{%
\includegraphics[width=0.7in]{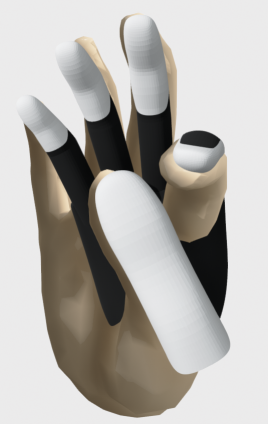}\label{fig:embodiment_interactive_mia_3}}%
\hfill
\subfloat[]{%
\includegraphics[width=0.7in]{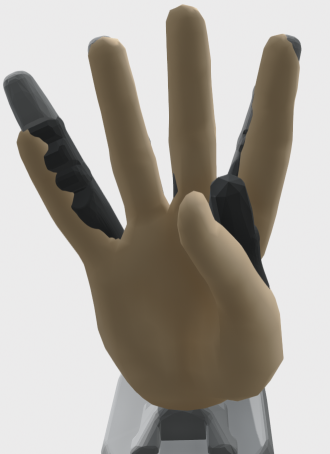}\label{fig:embodiment_interactive_shadow_1}}%
\hfill
\subfloat[]{%
\includegraphics[width=0.7in]{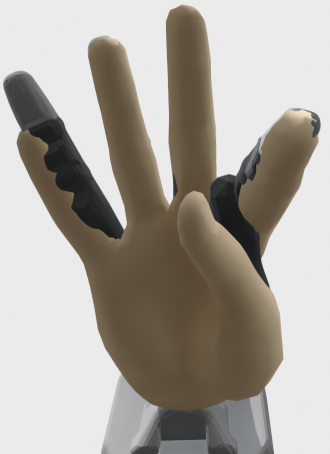}\label{fig:embodiment_interactive_shadow_2}}%
\hfill
\subfloat[]{%
\includegraphics[width=0.7in]{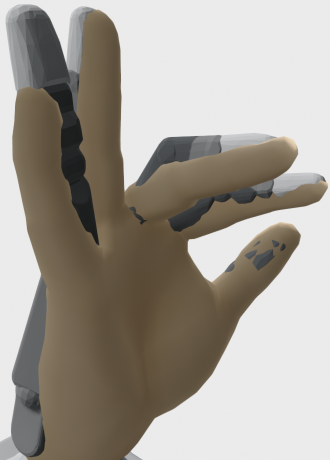}\label{fig:embodiment_interactive_shadow_3}}%
\hfill
\subfloat[]{%
\includegraphics[width=0.7in]{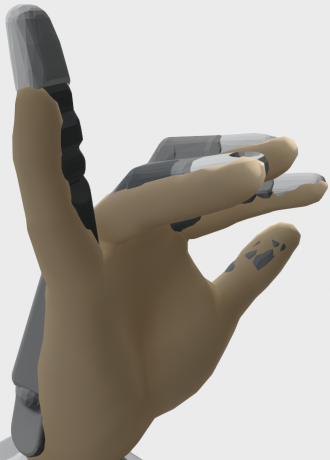}\label{fig:embodiment_interactive_shadow_4}}%
\hfill
\subfloat[]{%
\includegraphics[width=0.7in]{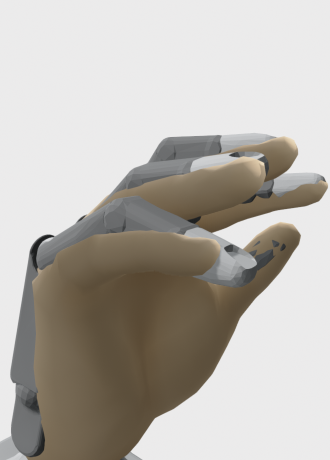}\label{fig:embodiment_interactive_shadow_5}}%
\hfill
\caption{Interactive embodiment mapping. MANO state and robotic hand after
embodiment mapping are displayed together. This visualization is based on Open3D's
visualizer \cite{Zhou2018} and pytransform3d \cite{Fabisch2019}.}
\label{fig:embodiment_interactive}
\end{figure}

Figure \ref{fig:embodiment_interactive} shows the result of an interactive
embodiment mapping. A GUI application was used to
set the 48 joint parameters of the MANO model. The embodiment mapping
determines joint angles of the robotic hand. Both the MANO mesh and the
configuration of the robotic hand are visualized.

Figure \ref{fig:embodiment_interactive} shows exemplary configurations
of the Mia hand (a~--~c) and configurations of the Shadow dexterous hand (d~--~h).
There are differences between the MANO mesh and the robotic hand
that the embodiment mapping cannot compensate for: the Mia hand is slightly
smaller than the MANO mesh so that the little finger cannot be aligned 
perfectly, and the little finger of the Shadow dexterous hand is longer than the
one of the MANO mesh. The Mia hand has only 4 DOF,
which results in a less accurate embodiment, in particular when the middle finger,
ring finger, and little finger have a different flexion as these move
jointly in the Mia hand. There are also differences that occur due to an
inadequate objective: the Shadow dexterous hand is able to minimize the
positional difference between the finger tips without having the correct
orientation (see Figure \ref{fig:embodiment_interactive_shadow_1}) and
as long as the tip positions are reached it does not matter whether the joint
angles are similar (e.g., see Figure \ref{fig:embodiment_interactive_shadow_5}).
As it will become apparent in Sections \ref{subsec:similarity} and
\ref{subsec:realtime}, the last point is the price that we pay to for
real-time control of a robotic hand.

\subsection{Similarity Between MANO and Robotic Hand}
\label{subsec:similarity}

\begin{figure}[bt]
\centering
\hfill
\subfloat[MANO.]{%
\includegraphics[height=1in]{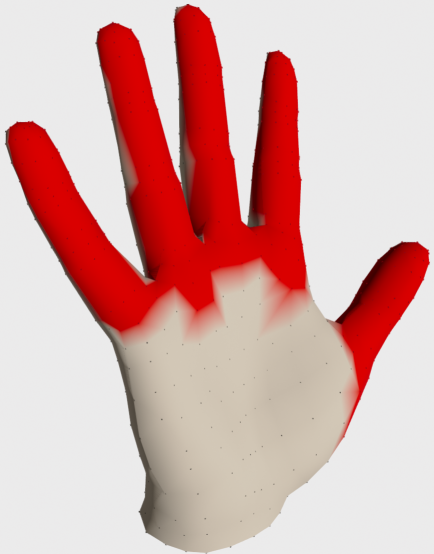}\label{fig:mano_contact_fingers}}
\hfill
\subfloat[Mia hand.]{%
\includegraphics[height=1in]{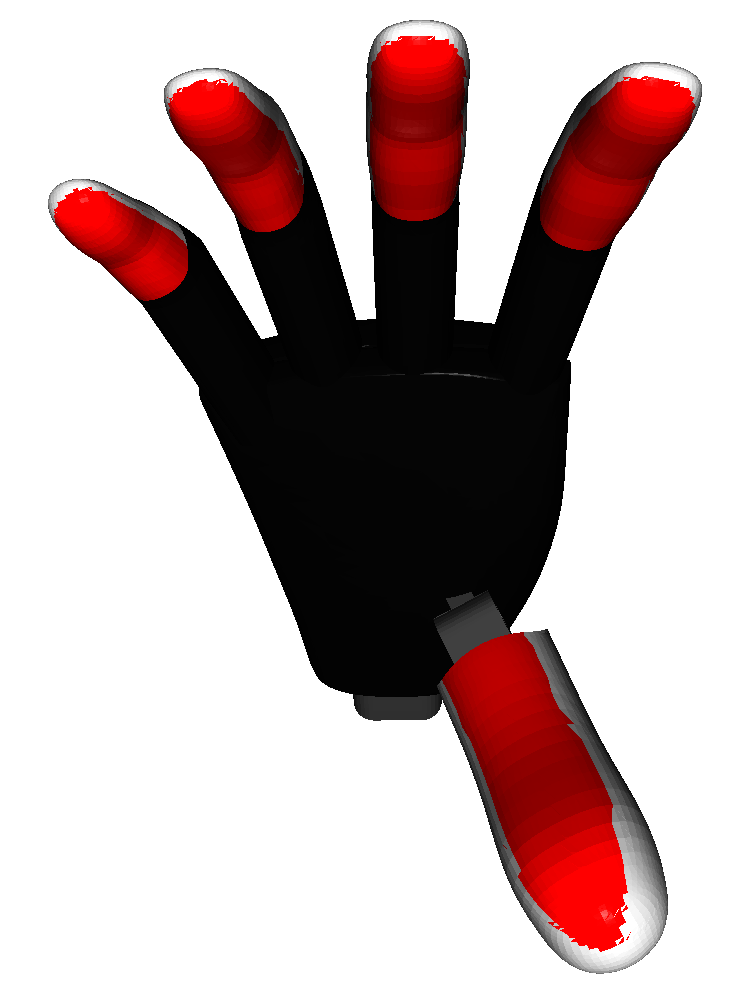}\label{fig:mia_contact}}
\hfill
\subfloat[Shadow hand.]{%
\makebox(70,75){
\includegraphics[height=1in]{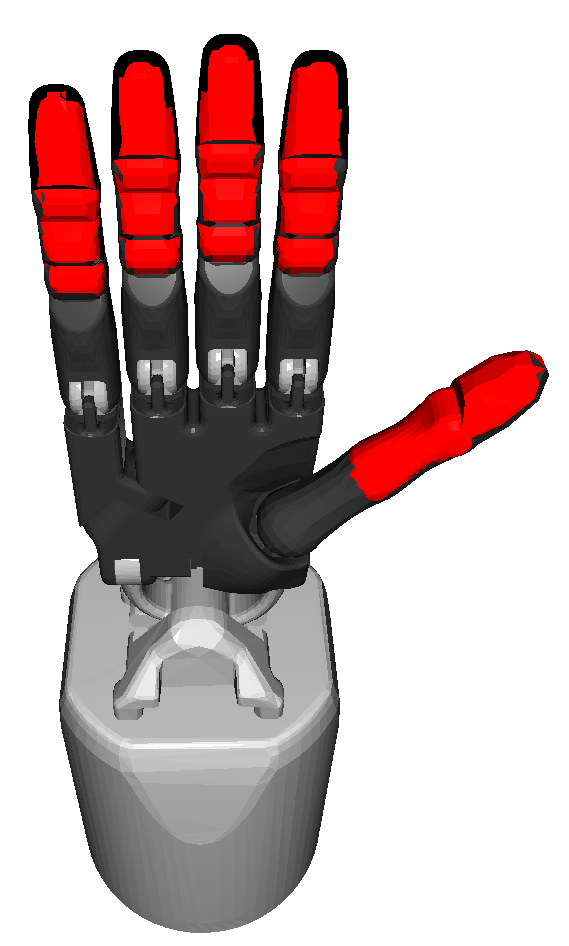}\label{fig:shadow_contact}}}
\hfill
\caption{Contact surfaces of the MANO model and the robotic hands are marked in red color.}
\label{fig:contact_surfaces}
\vskip -1em
\end{figure}

\begin{table}[bt]
\centering
\vskip 1em  % fix margin impositions
\begin{tabular}{lrrrrr}
\toprule
Hand & Thumb & Index & Middle & Ring & Little\\
\midrule
Mia hand              & 12.2 &  8.3 & 17.4 & 23.3 & 30.2\\
Shadow dexterous hand &  5.2 &  6.3 &  5.6 &  5.9 &  9.8\\
Robotiq 2F-140        & 28.6 & 13.4 &  - &  - & -\\
\bottomrule
\end{tabular}
\caption{Mean average distance (unit: mm) of contact surfaces.
One frame per demonstration (763) was selected. A simple
Robotiq gripper with 1 DOF is used as baseline.}
\label{tab:results}
\vskip -1em
\end{table}

We compare inner surfaces of fingers of the robotic hands to the MANO
mesh to evaluate the embodiment. Hence, we define the contact surfaces
of the MANO model and each robotic hand for each finger that we compare
(see Figure \ref{fig:contact_surfaces}).

For evaluation we draw 100 points from the contact surface of each finger
of the robotic hand by Poisson disk sampling \cite{Yuksel2015}, compute the
minimum distance to the closest triangle of the corresponding surface
of the MANO mesh per sample, and average these minimum distances over
all samples per finger. More precisely, we compute
$
\frac{1}{N} \sum_{i=1}^N \min_{j \in \{1, \ldots, M \}}
d(\boldsymbol{p}_i, T_j),
$
where $d(\boldsymbol{p}, T)$ is the distance between a point and a triangle
\cite{Ericson2004}, $\boldsymbol{p}_i$
are points on the contact surface of the robotic finger,
$T_j$ are triangles of the corresponding
finger of MANO, $N$ is the number of samples from
%the contact surface of
the robotic finger, and $M$ is the number of triangles
on the contact surface of MANO.
The result is an average distance between the two surfaces.
Since the computation is considerably slower than the embodiment,
we do it for selected cases only.

Table \ref{tab:results} shows that more
DOF enable the embodiment mapping to fit desired configurations
more closely. The Mia hand, e.g., has problems with fitting the
middle, ring and little fingers because they are controlled by the same motor.
% Furthermore, the Mia hand is slightly smaller than the MANO mesh in the used
% configuration, as seen in Figure \ref{fig:embodiment_interactive}. With this
% transformation between MANO and the base of the robotic hand, however, it is
% easier to fit the thumb and the index finger accurately.

\subsection{Transfer to Simulation}

We use PyBullet \cite{Coumans2021}, one of the few physics engines that support
robots and deformable objects, to verify physical plausibility of transferred
motions.
Since setting up realistic simulation environments and modeling deformable
objects that have complex shapes is difficult, we focus on the
tasks of grasping an insole and a small pillow.
We model them as homogeneous objects with the stable Neo-Hookean model
\cite{Smith2018} for hyperelastic material. For
the insole we set Young's modulus to $E = \SI{100}{\kilo\pascal}$ and
Poisson's ratio to $\nu = 0.2$. For the pillow we set
$E = \SI{10}{\kilo\pascal}$ and $\nu = 0.2$.

\begin{figure}[tb]
\centering
\vskip 1em
\subfloat[Insole and Mia hand.]{%
\includegraphics[height=0.8in]{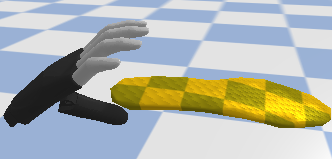}\label{fig:sim_insole}}
\hfill
\subfloat[Pillow and Shadow hands.]{%
\includegraphics[height=0.8in]{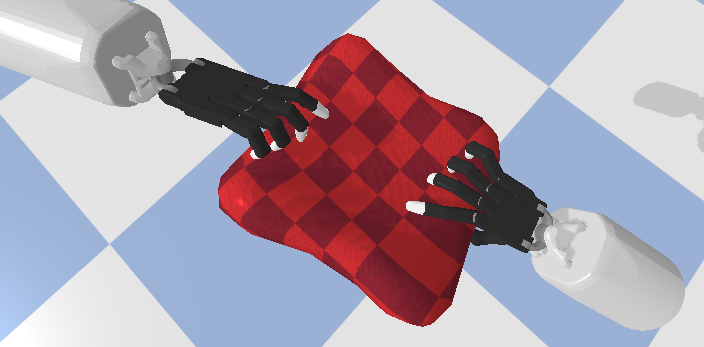}\label{fig:sim_pillow}}
\caption{Simulation with floating hands.}
\label{fig:sim}
\vskip -1em
\end{figure}

We test whether the object can be held after the execution of each
grasp by simulating the effect of gravity.
The objects float initially (see Figure \ref{fig:sim}).
%This is achieved by fixing a subset of nodes of the object mesh at their
%positions during grasp execution.
After each completed grasp, we evaluate its success by
%removing the constraints from the simulation,
allowing the object to fall from gravitational
force. We continue the simulation for two seconds
and measure whether the object is still in the hand.
To exclude problems of reachability, we simulate only floating
hands without a robotic arm.
Since the pillow is much larger than both hands, the best strategy is to grasp
it with two hands. Hence, we execute the same demonstrated grasp with two
hands, where one trajectory is rotated 180 degrees around the axis pointing
up in the middle of the pillow, which works because the pillow is symmetric.
Otherwise a second human hand would have to be recorded.

Table \ref{tab:sim_results} shows the success rate of the embodiment mapping
for each combination of task and hand. Considering morphological differences
between the human hand and the robotic hands 100\% success rate is hardly achievable.
Despite resembling MANO states more faithfully, the Shadow hand does not
perform better than the Mia hand in these tasks. We attribute this to the fact that it
is not necessary to have many DOF to match the recorded human cylindrical and pinch grasps.
In fact, the geometry of the Mia hand is better suited to grasp these two
objects firmly than the Shadow hand, mainly due to its big thumb.
Note that we expect grasping to work better when force sensors are used as
feedback. Thus, it is best to use the transferred motion in
combination with a controller that could, e.g., be generated through reinforcement
learning.
Modeling contact and friction of deformable objects in simulation is difficult.
Hence, we perform experiments on the real system to check whether we can trust
results from simulation.

\begin{table}[tb]
\centering
\begin{tabular}{lclr}
\toprule
Object & Samples & Hand & Success Rate\\
\midrule
\multirow{2}{*}{Insole} & \multirow{2}{*}{213} & Mia & 71.4\%\\
 & & Shadow & 40.4\%\\
\midrule
\multirow{2}{*}{Small pillow} & \multirow{2}{*}{224} & Mia & 92.8\%\\
 & & Shadow & 65.2\%\\
\bottomrule
\end{tabular}
\caption{Success rates of simulated grasps.}
\label{tab:sim_results}
\vskip -1em
\end{table}

\subsection{Transfer to Real System}

\subsubsection{Methods}
The trajectories generated by the embodiment were tested in a real robot. The set-up comprises a robot arm (UR10e, Universal Robots), a 6-axis force-torque sensor (HEX-E v2, OnRobot) mounted at the wrist of the robot arm, and an anthropomorphic artificial hand (MIA hand, Prensilia). The target object is a deformable insole bending at the edge of a table and the grasp point is outside of the table. During the tests, the insole was positioned in the same location by means of a mask. The arm and the hand were controlled at a frequency of 100 Hz and 20 Hz, respectively. The data from the embodiment mapping were used to control the robot via a multi-node ROS environment. The experiment included a subset of 80 trajectories differing for grasp location (i.e., 40 tip and 40 heel), but characterized by the same grasp type (i.e., cylindrical). The experiment comprised of two experimental conditions. The first (Coordinated Trajectories - CT), is aimed at assessing the capacity of the embodiment to successfully transfer coordinated motions. Thus, in this session, the reaching motion of the arm and the grasping action of the hand were controlled in a coordinated fashion as computed by the embodiment. The second (Sequential Trajectories - ST) sought to assess the success rate when the reaching motion of the arm is accomplished before the grasping action of the hand.
For each trial, the robot executed the trajectory of the arm and the joint trajectories of fingers to grasp the insole.  At the end, the robot moved toward the human operator through a predefined trajectory. The grasp was judged successful if the object did not fall during the lifting and transporting phases (without changing the object orientation as for the simulation). Finally, if the grasping action of the robot succeeded, the human operator pulled the object out of the hand along the direction of the long fingers. In this phase, we recorded the magnitude of the force at the wrist of the robot and used this data to indirectly evaluate the stability of the grasp.

\subsubsection{Results and discussion}
Results of the experiments are summarized in Table \ref{tab:exp_results}. Overall, the success rate is 83.8\% (67 out of 80 trajectories) for the CT condition, and 62.5\% (50 out of 80 trajectories) for the ST condition. This trend is also confirmed by looking at the performances for the different grasp locations. These results show that the embodiment mapping preserves the benefits of the human motor coordination in the action of grasping. Among the different grasp locations, the heel has the highest success rate (95\% and 75\% for the CT and ST conditions, respectively).
Successful trajectories can lead to stable or slightly stable grasps, and we used the force recorded during the pull-out phase to discriminate among these two classes. A grasp was judged stable if the maximum of the magnitude of the force is greater than a threshold of 2.8 N (this value was set based on the performance of the sensor used). Overall, the success rate of stable grasps is 72.5\% (58 out of 80 trajectories) for the CT condition. This result is aligned with the output of the simulation, being 70\% considering this subset of 80 trajectories (Table \ref{tab:exp_results}). 
%Small deviations between simulation and reality can be explained by simplified models of deformable objects and friction that were used in simulation.

\begin{table}[tb]
\centering
\begin{tabular}{ccrcc}
\toprule[.1em]
			&  			    & 	   	 & Success rate    & Success (trajectories) \#   \\
 \midrule[.1em]

 \multirow{9}*{\rotatebox{90}{\parbox{3.2cm}{\centering\textbf{Coordinated\\ Trajectories}}}} & \multirow{3}*{Tip} & All    & 72.5\%    & 29 (40)    \\
 %\cline{3-5} 
 &           &  Stable    & 62.5\%   & 25 (40)    \\
 %\cline{3-5} 
 &           &   {\it Simulation} & 62.5\%    & 25 (40)    \\
	\cmidrule{2-5} & \multirow{3}*{Heel} & All    & 95.0\%    & 38 (40)    \\
 %\cline{3-5} 
 &           &  Stable    & 82.5\%   & 33 (40)    \\
 %\cline{3-5} 
 &           &   {\it Simulation} & 77.5\%    & 31 (40)    \\
\cmidrule{2-5} & \multirow{3}*{Overall} & All    & 83.8\%    & 67 (80)    \\
% \cline{3-5} 
 &           &  Stable    & 72.5\%   & 58 (80)    \\
% \cline{3-5} 
 &           &   {\it Simulation} & 70.0\%    & 56 (80)    \\

\midrule[.1em]

 \multirow{6}*{\rotatebox{90}{\parbox{2.2cm}{\centering\textbf{Sequential\\ Trajectories}}}} & \multirow{2}*{Tip} & All    & 50.0\%    & 20 (40)    \\
 %\cline{3-5} 
 &           &  Stable    & 42.5\%   & 17 (40)    \\
 %\cline{3-5} &           &   {\it Simulation} & 65\%    & 26 (40)    \\
	\cmidrule{2-5} & \multirow{2}*{Heel} & All    & 75.0\%    & 30 (40)    \\
% \cline{3-5} 
 &           &  Stable    & 62.5\%   & 25 (40)    \\
 %\cline{3-5} &           &   {\it Simulation} & 75\%    & 30 (40)    \\
\cmidrule{2-5} & \multirow{2}*{Overall} & All    & 62.5\%    & 50 (80)    \\
 %\cline{3-5} 
 &           &  Stable    & 53.8\%   & 43 (80)    \\
 %\cline{3-5} &           &   {\it Simulation} & 70\%    & 56 (80)    \\

\bottomrule[.1em]
\end{tabular}
\caption{Success rates and stability of the grasps for the set of 80 trajectories executed in coordinated and sequential fashion.}
\label{tab:exp_results}
\vskip -1em
\end{table}

\subsection{Real-Time Control Capabilities}
\label{subsec:realtime}

% \begin{table*}[bt]
% \centering
% \vskip 1em  % fix margin impositions
% \begin{tabular}{lrccccccccccc}
% \toprule
% & & \multicolumn{3}{c}{Record mapping} & \multicolumn{3}{c}{Embodiment mapping (Mia)} & \multicolumn{3}{c}{Embodiment mapping (Shadow)}\\
% \cmidrule{3-11}
% & & \multicolumn{9}{c}{Frequency in Hz}\\
% \cmidrule{3-11}
% Dataset & Frames & average & min & max & average & min & max & average & min & max\\
% \midrule
% Grasp insole & 42,285 & 99.7 & 5.1 & 123.2 & 228.0 & 15.0 & 328.6 & 178.7 & 0.7 & 202.2\\
% Insert insole & 3,878 & 82.9 & 6.2 & 112.2 & 280.5 & 73.2 & 313.3 & 162.2 & 8.5 & 201.1\\
% Grasp small pillow & 31,020 & 65.8 & 5.9 & 126.1 & 261.1 & 102.2 & 322.6 & 136.7 & 9.0 & 204.6\\ % record std. 80.2 embodiment std. 1063.8
% Grasp big pillow & 6,945 & 58.7 & 5.6 & 115.1 & 268.7 & 97.0 & 322.4 & 131.2 & 8.3 & 203.5\\
% Grasp eletronic component & 9,459 & 84.1 & 7.2 & 129.7 & 292.2 & 131.0 & 332.5 & 169.1 & 8.6 & 207.8 \\
% Assemble eletronic components & 16,984 & 95.2 & 7.7 & 121.3 & 293.9 & 123.1 & 328.7 & 180.7 & 10.5 & 212.3\\
% Flip pages & 15,484 & 63.4 & 9.1 & 124.1 & 263.1 & 108.5 & 341.7 & 128.0 & 0.7 & 218.4 \\
% Insert passport in box & 6,669 & 58.2 & 3.0 & 109.6 & 158.6 & 16.5 & 304.3 & 142.7 & 8.1 & 202.9 \\
% \bottomrule
% \end{tabular}
% \caption{Evaluation of speed. Results are statistics of each frame of each
% demonstration of the corresponding skill. All computations are done by a single
% core of an AMD Ryzen 7 2700 processor.}
% \label{tab:evaluation_real_time}
% \vskip -1em
% \end{table*}

\begin{table}[bt]
\centering
\vskip 1em  % fix margin impositions
\begin{tabular}{lr|cc|cc|cc}
\toprule
\multirow{5}{*}{\rotatebox{90}{Task no.}} & & \multicolumn{6}{c}{Record/Embodiment Mapping Frequency in Hz}\\
\cmidrule{3-8}
& & \multicolumn{2}{c|}{MANO} & \multicolumn{2}{c|}{Mia} & \multicolumn{2}{c}{Shadow}\\
\cmidrule{3-8}
& Frames & mean & min & mean & min & mean & min\\
\midrule
1 & 42,285 & 99.7 & 5.1 & 228.0 & 15.0 & 178.7 & 0.7 \\
%2 & 3,878 & 82.9 & 6.2 & 280.5 & 73.2 & 162.2 & 8.5 \\
3 & 31,020 & 65.8 & 5.9 & 261.1 & 102.2 & 136.7 & 9.0 \\
%4 & 6,945 & 58.7 & 5.6 & 268.7 & 97.0 & 131.2 & 8.3 \\
%5 & 9,459 & 84.1 & 7.2 & 292.2 & 131.0 & 169.1 & 8.6 \\
6 & 16,984 & 95.2 & 7.7 & 293.9 & 123.1 & 180.7 & 10.5 \\
7 & 15,484 & 63.4 & 9.1 & 263.1 & 108.5 & 128.0 & 0.7 \\
%8 & 6,669 & 58.2 & 3.0 & 158.6 & 16.5 & 142.7 & 8.1 \\
\bottomrule
\end{tabular}
\caption{Evaluation of speed. Results are statistics of each frame of each
demonstration of the task (only tasks with $>$10,000 frames). Computations are done by one
core of an AMD Ryzen 7 2700 CPU. Task numbers refer to Table \ref{tab:datasets}.}
\label{tab:evaluation_real_time}
\vskip -1em
\end{table}

One intended use case of the embodiment mapping is to enable real-time
control of a robotic arm and hand through a motion capture system.
%This allows the operator to compensate for differences of their hand and the robotic hand during execution.
A limiting factor is the frequency at which we receive hand states
from the motion capture system, which is 100 Hz. We must generate commands
for the robotic hand from motion capture with a similar frequency.
Table \ref{tab:evaluation_real_time} shows the frequencies at which we are
able to compute record and embodiment mapping. While
the lowest frequencies for both mappings prevent real
time control even with a control frequency of \SI{20}{\hertz} for the Mia hand, the
average frequency of the embodiment mapping is well above the frequency at
which the motion capture system provides measurements.
The record mapping is often too slow for \SI{100}{\hertz}.
Nevertheless, it is possible to split hand pose estimation, which can be done
at a high frequency, and estimation of finger configuration, which does not
need to be done at a high frequency to control the Mia hand.
% Nevertheless, in a setup with a Universal Robot that is controlled at 125 Hz
% and a Mia hand that is controlled at 20 Hz, we can compute the end-effector
% poses for the arm at a higher frequency, as this is not computationally
% demanding, and provide the computationally demanding finger configurations
% at a lower frequency so that real-time control is possible.

\section{CONCLUSIONS}

We show that MANO states can be obtained from MOCAP
in addition to the usual approach with one camera.
Using MOCAP allows for more complex, natural motions than kinesthetic teaching
and occlusions are less likely than with a single camera.
Furthermore, we introduce a modular framework to transfer human hand motions to two robotic
hands with varying complexity through a configurable embodiment mapping that is
fast enough for complex robotic hands.
%Advantages of using a motion capture system include high-precision, low-variance hand pose estimates and accurate measurements of the finger-tips' cartesian positions compared to estimates from single camera images.
Even without feedback, most embodied trajectories could successfully solve
the task of grasping simulated insoles and pillows. Real results are
aligned with simulation.
Experiments also show that coordination between hand pose and finger movements
is more effective than sequential execution, which emphasizes the relevance of
our approach.
However, results could be improved with, e.g., reinforcement learning.
On the one hand this would considerably reduce the necessary exploration for
reinforcement learning and on the other hand it would make transferred motions more robust.
%The evaluation shows promising results in terms of similarity of transferred
%hand states as well as enough speed for real-time performance, which enables
%teleoperation of robotic hands.


%%%%%%%%%%%%%%%%%%%%%%%%%%%%%%%%%%%%%%%%%%%%%%%%%%%%%%%%%%%%%%%%%%%%%%%%%%%%%%%%

%%%%%%%%%%%%%%%%%%%%%%%%%%%%%%%%%%%%%%%%%%%%%%%%%%%%%%%%%%%%%%%%%%%%%%%%%%%%%%%%

%%%%%%%%%%%%%%%%%%%%%%%%%%%%%%%%%%%%%%%%%%%%%%%%%%%%%%%%%%%%%%%%%%%%%%%%%%%%%%%%
%\section*{APPENDIX}


\section*{ETHICS APPROVAL}

Experimental protocols were approved by the ethics committee of the University
of Bremen. Written informed consent was obtained from all participants.

\section*{ACKNOWLEDGMENT}

We thank Oscar Lima, Andrea Burani, and Francesca Cini for the URDF of the Mia hand
and Lisa Gutzeit for her feedback on the manuscript.
The motion capture setup was developed in collaboration with Lisa Gutzeit,
supported by a grant from the German Federal Ministry for Economic Affairs and Energy
(BMWi, FKZ 50 RA 2023).

%%%%%%%%%%%%%%%%%%%%%%%%%%%%%%%%%%%%%%%%%%%%%%%%%%%%%%%%%%%%%%%%%%%%%%%%%%%%%%%%

\bibliographystyle{plain}
%\bibliography{literature}

\end{document}